\RequirePackage{fix-cm}
\documentclass[smallextended]{svjour3}       

\smartqed  
\usepackage{graphicx}
\usepackage{amssymb}

\usepackage{amsmath}
\usepackage{algorithm}
\usepackage[noend]{algpseudocode}
\usepackage{xcolor}

\usepackage{hyperref}
\usepackage{multirow}

\usepackage{lineno}

\DeclareMathOperator*{\argmin}{arg\,min}

\makeatletter
\def\BState{\State\hskip-\ALG@thistlm}
\makeatother

\begin{document}

\title{How deep should be the depth of convolutional neural networks: a backyard dog case study}

\author{Alexander N. Gorban \and Evgeny M. Mirkes \and Ivan Y. Tyukin}

\institute{Alexander N. Gorban \at
              University of Leicester, Leicester LE1 7RH, UK \\ and Lobachevsky State University of Nizhny Novgorod, Prospekt Ganarina 23, 603950,  Nizhny Novgorod, Russian Federation \\
               \email{a.n.gorban@le.ac.uk}           
                     \and
           Evgeny M. Mirkes \at
              University of Leicester, Leicester LE1 7RH, UK \\ and Lobachevsky State University of Nizhny Novgorod, Prospekt Ganarina 23, 603950,  Nizhny Novgorod, Russian Federation \\
              \email{em322@le.ac.uk}
                        \and
            Ivan Y. Tyukin \at
            University of Leicester, Leicester LE1 7RH, UK \\ Tel.: +44 116 2525106,  \\ and Lobachevsky State University of Nizhny Novgorod, Prospekt Ganarina 23, 603950,  Nizhny Novgorod, Russian Federation \\ and Saint-Petersburg State Electrotechnical University (LETI), Prof. Popova 5, Saint-Petersburg, Russian Federation
              \email{I.Tyukin@le.ac.uk}
}

\maketitle

\newpage

\section*{Structured abstract}

\subsection*{Background}

The work concerns the problem of reducing a {pre-trained} deep neuronal network  to a smaller network, with just few layers, {whilst retaining the network's functionality on a given task}. {In this particular case study, we are focusing on the networks developed for the purposes of face recognition.}

\subsection*{Methods}

{The proposed approach is motivated by the observation that the aim to deliver the highest accuracy possible in the broadest range of operational conditions, which many deep neural networks models strive to achieve, may not necessarily be always needed, desired, or even achievable due to the lack of data or technical constraints. In relation to the face recognition problem, we formulated an example of such a usecase,  the `backyard dog' problem. The `backyard dog', implemented by a lean network, should correctly identify members from a limited group of individuals, a `family', and should distinguish between them. At the same time, the network must produce an alarm to an image of an individual who is not in a member of the family, i.e. a `stranger'. To produce such a lean network, we propose a network shallowing algorithm. The algorithm takes an existing deep learning model on its input and outputs a shallowed version of the model. The algorithm is non-iterative and is based on the Advanced Supervised Principal Component Analysis. Performance of the algorithm is assessed in exhaustive numerical experiments.}

\subsection*{Results}

Our experiments revealed that {in the above usecase, the `backyard dog' problem,} the method is capable of drastically reducing the depth of deep learning neural networks, albeit at the cost of mild performance deterioration.

\subsection*{Conclusion}

In this work we proposed a simple non-iterative method for shallowing down pre-trained deep convolutional networks. The method is generic in the sense that it applies to a broad class of feed-forward networks, and is based on the Advanced Supervise Principal Component Analysis. The method enables {generation of families} of smaller-size shallower specialized networks tuned for specific operational conditions and tasks from a single larger and more universal legacy network.

\subsection*{Keywords}

Non-iterative learning \and Principal Component Analysis \and Convolutional Neural Networks

\newpage

\section{Introduction}
\label{intro}

{With the explosive pace of progress in computing, availability of cloud resources and open-source dedicated software frameworks, current Artificial Intelligence (AI) systems are now capable of spotting minute patterns in large data sets and may outperform humans and early-generation AIs in highly complicated cognitive tasks including object detection \cite{Krizhevsky2012}, medical diagnosis \cite{Huiying2019},  face and facial expression recognition \cite{Xiao2019},\cite{Ranjan2018}. At the centre of these successes are deep neural networks and deep learning technology  \cite{Zhong-Qiu2019}, \cite{LeCun2015}.}

{Despite this, several fundamental challenges remain which constrain and impede further progress. In the context of  face recognition \cite{Ranjan2018}, these  include the need for  larger volumes of high-resolution and balanced training and validation data  as well as the inevitable presence of hardware constraints limiting training and deployment of large models. Consequences of imbalanced training and testing data may have significant performance implications. At the same time, hardware limitations, such as memory constraints, restrict adoption, development, and spread of technology. These challenges constitute fundamental obstacles for creation of universal data-driven AI systems, including for face recognition.}

The challenge of overcoming hardware limitations whilst maintaining functionality of the underlying AI received significant attention in the literature. Heuristic definition of an efficient neural network was proposed in 1993: delivery of maximal performance (or skills) with minimal number of connections (parameters) \cite{Gordienko1993}. Various algorithms of neural networks optimization were proposed in the beginning of 1990s \cite{Gorban1990,Hassibi1993}.  MobileNet \cite{MobileNet}, SqueezeNet \cite{SqueezeNet}, DeepRebirth \cite{LiWangKong2017}, and EfficientNets \cite{EfficientNet}  are more recent examples of the approaches in this direction. {Notwithstanding, however, the need for developing generic and flexible universal systems for a wide spectrum of tasks and conditions,  there is a range of practical problems in which such universality may not be needed or required. These tasks may require smaller volumes of data and could be deployed on cheaper and accessible hardware. It is hence imperative that these tasks are identified and investigated, both computationally and analytically.}

{In this paper we present and formally define such a task in the remit of face recognition: the `backyard dog' problem. The task, on the one hand, appears to be a close relative of the standard face recognition problem. On the other, it is more relaxed which enables us to lift limitations associated with the availability of data and computational resources. For this task, we propose a technology and an algorithm for constructing a family of the `backyard dog' networks derived from larger pre-trained legacy convolutional neural nets (CNN). The idea to exploit existing pre-trained networks is well-known in the face recognition literature  \cite{Simonyan2015},  \cite{Parkin2015}, \cite{Schroff2015}, \cite{Taigman2014}, \cite{Guoqiang2018}. Our algorithm shares some similarity with \cite{Guoqiang2018} in that it exploits existing parts of the legacy system and uses them in a dedicated post-processing step. In our case, however, we apply these steps methodically across all layers; at the post-processing step we employ advanced supervised Principal Component Analysis (PCA)  \cite{Mirkes2016}, \cite{KorenCarmel2004} rather than conventional PCA, and do not use support vector machines.

Implementation of the technology and performance of the algorithm is illustrated with a particular network architecture,  VGG net \cite{Parkin2015}, and implemented on two computational platforms.  The first platform  was Raspberry Pi 3B with Broadcom BCM2387 chipset, 64 bit CPU 1.2GHz Quad-Core ARM Cortex-A53 and 1 GiB memory with OS Raspbian Jessie. We will refer to it as `Pi'. The second platform was HP EliteBook laptop with Intel Core i7-840QM (4 x 1.86 GHz)  CPU and 8GiB of memory with OS Windows 7. We refer to this platform as `Laptop'. In view of Pi3 memory limitations (1 GiB), we  required that the `backyard dog' occupies no more than than  300 MiB. The overall workflow, however, is generic and should transfer well to other models and platforms.}

The manuscript is organized as follows: in Section \ref{review} we review the conventional face recognition problem, formulate the `backyard dog' problem, assess several popular deep network architectures, and select a test-bed architecture for implementation; Section \ref{BDresults} describes the proposed shallowing technology for creation of the `backyard dog' nets and illustrates it with an example; Section \ref{Conclusion}  concludes the paper.

\section{Preliminaries and problem formulation}
\label{review}

Face recognition is arguably among the hardest technical and computational problems. If posed as a conventional multi-class classification problem, it is ill-defined as acquiring samples from all classes, i.e. all identifies, is  hardly possible. Therefore, state-of-the-art modern face recognition systems do not approach it as the multi-class classification problem. Not at least at the stage of deployment. These systems are often asked to answer another question: whether two given images correspond  to the same person or not.

The common idea is: map these images into a `feature space' with some metric (or a similarity measure) $\rho$. The system is then trained to ensure that if $x$ and $y$ are images corresponding to the same person then, for some $\varepsilon>0$, $\rho(x,y)<\varepsilon$, and $\rho(x,y)>\varepsilon$ otherwise. At the decision stage, if $\rho(x,y)<\varepsilon$ then $x,y$ represent the same person, and if $\rho(x,y)>\varepsilon$ then they belong to different identities.  The problem with these generic systems is that validation and performance quantification for such systems is challenging; they must work well for all persons and images, including for identities  these system never seen before.


It is thus hardly surprising that reports about performance of neural networks in face recognition tasks are often over-optimistic, with the accuracy of 98\% and above \cite{Parkin2015}, \cite{Schroff2015}, \cite{Taigman2014} demonstrated on few benchmark sets. There is a mounting evidence that the training set bias, often present in  face recognition datasets,  leads to deteriorated performance in real-life applications \cite{bias2018}. If we use a human as a benchmark, trained experts make 20\% mistakes on the faces they have never seen before \cite{WhiteErrorRate2015}. Similar performance figures have been reported for  modern face recognition systems  when they assessed identities from populations that were underrepresented  in the training data \cite{bias2018}. Of course, we must always strive to achieve most ambitious goals, and the grand face recognition challenge is not an exception. Yet, in a broad range of practical situations, generality of the classical face recognition problem is not always needed or desired.

In what follows we propose a relaxation of the face recognition problem that is a significantly better defined and is closer to the standard multi-class problem with {\it known classes}. We call this problem the `backyard dog' problem of which the specification is provided below.

\begin{description}
\item[{\it The `backyard dog' problem (Task)}]  {Consider a limited group of individuals, referred to as a `family members' (FM) or `friends'. Individuals who are not members of the family are referred to as `strangers'. A face recognition system, `the backyard dog', should (i) separate images of friends form that of strangers} and, at the same time (ii) should distinguish members of the family from each other (identity verification).

More formally, if $q$ is an image of a person $p$, and $Net$ is a `backyard dog' net, then $Net(q)$ must return an identity class of $q$ if $p\in FM$ and a label indicating the class of `strangers' if $p\notin FM$.
\item[{\it The `backyard dog' problem (Constraints)}] The `backyard dog' must be generate decisions within a given time frame on a given hardware and occupy no more than a given volume of RAM.
\end{description}

{The difference between the `backyard dog'  problem and the traditional face recognition task is two-fold. First, the `backyard dog'  should be able to reliably discriminate between a relatively small set of {\it known} identity classes (members of the family in the `backyard dog' problem)  as opposed to  the challenge of reliable discrimination between pairs of images from a huge set of {\it unknown} identity classes (traditional face recognition setting). This is a significant relaxation as  existing collections of training data used to develop models for face recognition (see Table~\ref{Databases})  are several orders of magnitude smaller than 7.6 billion of the total world population \cite{Population}.  In addition, the `backyard dog' must separate a relatively small set of {\it known} friends from the huge but {\it unknown} set of potential strangers.} The latter task is still challenging but its is difficulty is largely reduced relative to the original face recognition problem in that it is now a binary classification problem.

In the next sections we will present a solution to the `backyard dog' problem in which we will take advantage of the availability  of a pre-trained deep legacy system. Before, however, presenting the solution let us first select a  candidate for  a legacy system that would allow us to illustrate the concept better. For this purpose, below we review and assess some of the well-known existing  system.

\subsection{VGG}

Oxford Visual Geometry Group (and hence the name VGG) published their version of CNN for face recognition in \cite{Parkin2015}. We call this network VGGCNN \cite{VGG2014}. The network was trained on a database containing facial images of  2622 different identities. Small modification of this network allows to compare two images and decide whether these two images correspond to the same person or not.

VGGCNN contains about 144M of weights. The recommended test procedure is as follows \cite{Parkin2015}:
\begin{enumerate}
  \item Scale detected face to three sizes: 256, 384, and 512.
  \item Crop a 224x224 fragment from each corner and from the centre of the scaled image.
  \item Apply horizontal flip to crops.
\end{enumerate}
Therefore, to test one face (one input image), one has to process 30 pre-processed images: $3({\rm scales})\times 5({\rm crops})\times 2 ({\rm flip})=30({\rm images})$.

Processing one image in the MatLab implementation \cite{MatConvNet} on our Laptop took approximately 0.7s.  TensorFlow implementation of  the same required circa 7.3s.

\subsection{FaceNet}

Several CNNs with different architectures have been associated with the name  \cite{Schroff2015}:
\begin{itemize}
\item NN1 with images 220x220, 140M of weights and 1.6B FLOP,
\item NN2  with images 224x224, 7.5M of weights and 1.5B FLOP,
\item NN3 with images 160x160, 7.5M of weights and 0.744B FLOP,
\item NN4 with images 96x96, 7.5M of weights and 0.285B FLOP.
\end{itemize}
Here, FLOP stays for Floating Point Operations per image processing. The testing procedure for FaceNet uses one network evaluation per image.

\subsection{DeepFace}

FaceBook \cite{Taigman2014} proposed DeepFace architecture which,  similarly to VGG face, is initially trained within a multi-class setting. At the evaluation stage,  two replicas of the trained CNN  assess a pair of images and produce their corresponding feature vectors.  These are then passed into a separate network  implementing the predicate `The same person/Different persons'.

\subsection{Datasets}
A comparison of the different datasets used to train the above networks is presented in Table~\ref{Databases}. We can see that the dataset used to develop VGG net is apparently the largest, except for the  datasets used by Google, Facebook, or Baidu, which are not publicly available.

\begin{table}

\caption{Comparison of the datasets used to develop face recognition systems: (the table is presented in \cite{Parkin2015})}\label{Databases}
{\footnotesize
\begin{tabular}{|l|r|r|l|}
\hline
\multicolumn{1}{|c|}{Dataset} & \multicolumn{1}{c|}{Identities} & \multicolumn{1}{c|}{Images}& \multicolumn{1}{c|}{Link} \\\hline
LFW&5,749&13,233&\url{http://vis-www.cs.umass.edu/lfw/#download}\\\hline
WDRef \cite{Chen2012}&2,995&99,773& N/A\\\hline
CelebFaces \cite{Sun2014}&10,177&202,599& N/A\\\hline
VGG \cite{Parkin2015}&2,622&2.6M &\url{http://www.robots.ox.ac.uk/~vgg/data/vgg_face/}\\\hline
FaceBook \cite{Taigman2014}&4,030&4.4M& N/A\\\hline
Google \cite{Schroff2015}&8M&200M& N/A\\\hline
\end{tabular}}
\end{table}

\subsection{Comparison of VGGCNN, FaceNet and DeepFace}

\begin{figure}[ht]
\centering\includegraphics[width=0.8\linewidth]{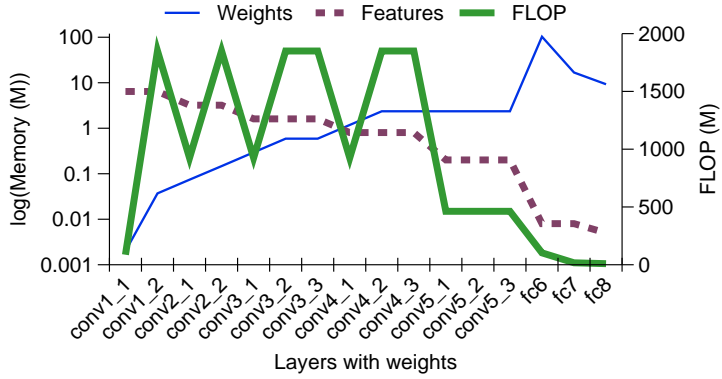}
\caption{Distribution of numbers of weights, features and required FLOP along VGG16 network from the deepest layers (left) to outputs (right)} \label{distrVGG}
\end{figure}

\begin{figure}[ht]
\centering\includegraphics[width=0.8\linewidth]{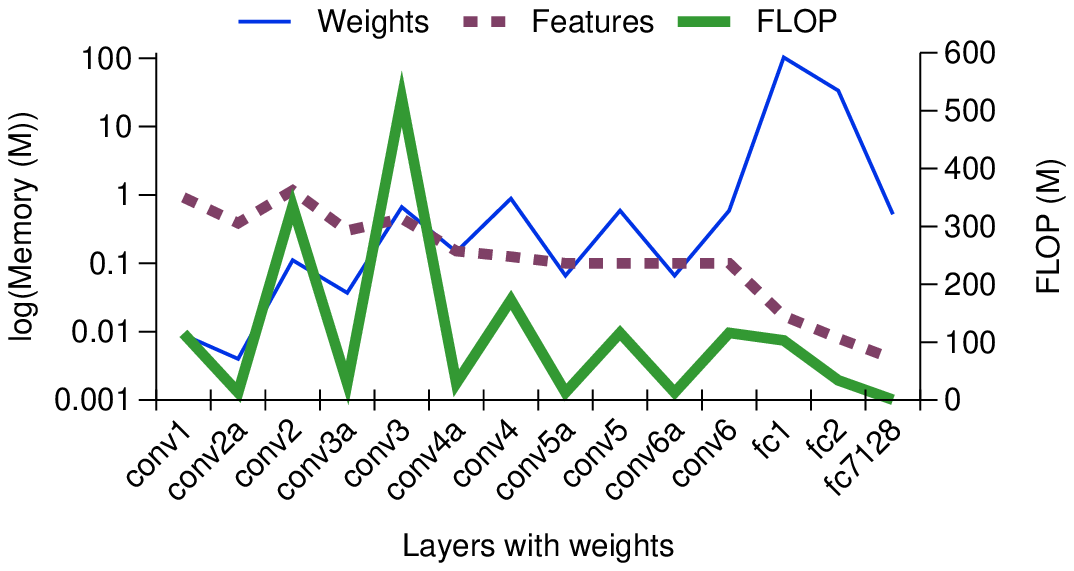}
\caption{Distribution of number of weights, features and required FLOP along NN1 network from the deepest layers (left) to outputs (right)} \label{distrNN1}
\end{figure}

\begin{figure}[ht]
\centering\includegraphics[width=0.8\linewidth]{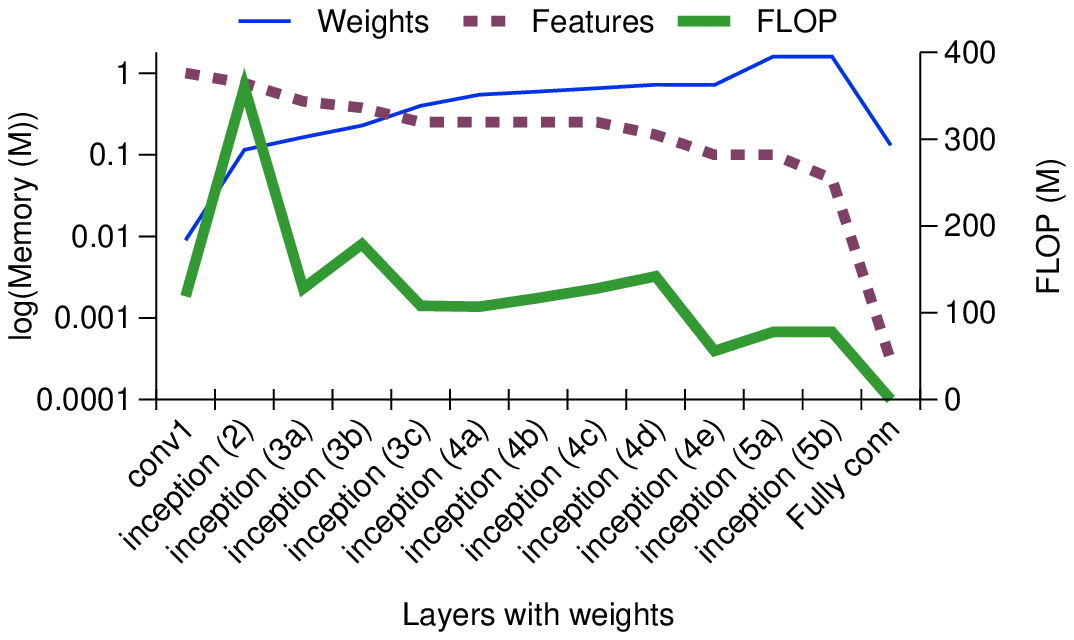}
\caption{Distribution of number of weights, features and required FLOP along NN2 network from the deepest layers (left) to outputs (right)} \label{distrNN2}
\end{figure}

\begin{figure}[ht]
\centering\includegraphics[width=0.8\linewidth]{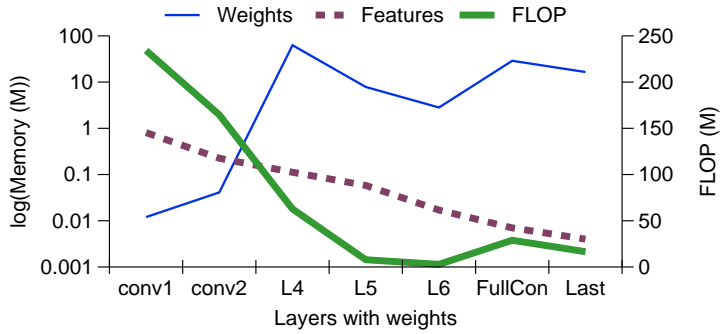}
\caption{Distribution of number of weights, features and required FLOP along DeepFace network from the deepest layers (left) to outputs (right)} \label{distrDeepFace}
\end{figure}

In addition to the training datasets, we have also compared  the volumes of weights (in MiBs) and computational resources (in  FLOPs) associated with each of the above networks. We did not evaluate their parallel/GPU-optimized implementations since our aim was to derive `'backyard dog' nets suitable for single-core implementations on the Pi platform. Results of this comparison are summarized in Table~\ref{MemTimeComp}, \ref{MemTimeComp2}.  Distributions of weights, features and time needed to propagate an image through each network are shown in Figs.~\ref{distrVGG} -- \ref{distrDeepFace}. Fig.~\ref{distrVGG} -- Fig.~\ref{distrDeepFace} also show that the interpretation of the notion of a `deep' network varies  for different teams: from 6 layers with weights in DeepFace to 16 such layers in VGG16.

\begin{table}
\caption{Memory requirements and computation resources needed: `Weights' is the number of weights in the entire network in millions of weights; `Features' is the maximal number of signals, in millions, which are passed from one layer to the other in a given network.}\label{MemTimeComp}{\footnotesize
\begin{tabular}{|l|l|c|r|r|r|r|}
\hline
\multicolumn{1}{|c|}{Developer} & \multicolumn{1}{c|}{Family name} & \multicolumn{1}{c|}{Name}  & \multicolumn{1}{c|}{Weights}  & \multicolumn{1}{c|}{Features}  &\multicolumn{1}{c|}{FLOP}  & \multicolumn{1}{c|}{Image} \\
 & &  & \multicolumn{1}{c|}{(M)}  & \multicolumn{1}{c|}{(M)}  &\multicolumn{1}{c|}{(M)}  & \multicolumn{1}{c|}{size} \\\hline

VGG group&VGGCNN \cite{Parkin2015}&VGG16&144.0&6.4&15,475&224\\\hline
\multirow{4}{*}{Google}&\multirow{4}{*}{FaceNet \cite{Schroff2015}}&NN1&140.0&1.2&1,606&220\\ \cline{3-7}
&&NN2&7.5&2.0&1,600&224\\\cline{3-7}
&&NN3&7.5&NA&744&160\\\cline{3-7}
&&NN4&7.5&NA&285&96\\\hline
FaceBook&DeepFace \cite{Taigman2014}&DeepFace-align2D&118.0&0.8&805&152\\\hline
\end{tabular} }
\end{table}

\begin{table}
\caption{Computational time needed for passing one image through different networks. For MatLab (ML) and TensorFlow (TF) realisations of VGGCNN on the Laptop platform, time was measured explicitly. All other values were estimated using FLOP data shown in Table~\ref{MemTimeComp} and taking VGGCNN data as a reference. Values for the Pi platform were estimated on the basis of explicit measurements for the reduced network (so that it fits into the system's memory) and then scaled up proportionally.}\label{MemTimeComp2}{\footnotesize
\begin{tabular}{|l|l|c|r|r|r|r|}
\hline
\multicolumn{1}{|c|}{Developer} & \multicolumn{1}{c|}{Family name} & \multicolumn{1}{c|}{Name}  & \multicolumn{1}{c|}{Laptop}  & \multicolumn{1}{c|}{Laptop}  &\multicolumn{1}{c|}{Pi TF}  & \multicolumn{1}{c|}{Pi 1} \\
 & &  & \multicolumn{1}{c|}{ML} & \multicolumn{1}{c|}{TF} &  & \multicolumn{1}{c|}{core C++} \\\hline

VGG group&VGGCNN \cite{Parkin2015}&VGG16&0.695&4.723&75.301&65.909\\\hline
\multirow{4}{*}{Google}&\multirow{4}{*}{FaceNet \cite{Schroff2015}}&NN1&0.072&0.490&7.815&6.840\\ \cline{3-7}
&&NN2&0.072&0.488&7.786&6.815\\\cline{3-7}
&&NN3&0.033&0.227&3.620&3.169\\\cline{3-7}
&&NN4&0.013&0.087&1.387&1.214\\\hline
FaceBook&DeepFace \cite{Taigman2014}&DeepFace-align2D&0.036&0.246&3.917&3.429\\\hline
\end{tabular} }
\end{table}

According to Table \ref{MemTimeComp2}, a C++ implementation for the Pi platform is comparable in terms of time with the TensorFlow (TF) implementation. Nevertheless, we note that we did not have control over the TF implementation in terms of enforcing the single-core  operation. This may explain why single image processing times for the C++ and the TF implementations are so close.

In summary, we conclude that all these networks require at least 30MiB of RAM for weights ($7.5\times 4$ MiB) and 3.2MiB for features. Small networks (NN2-NN4) satisfy the imposed memory restrictions of 300MiB. Large networks like VGG16, NN1 or DeepFace require more than 100M of weights or 400MiB and hence do not conform the this requirement. Time-wise, all candidate networks needed more than 1.2 seconds, with the VGGCNN requiring more than a minute on the Pi platform to process  an image.

Having done this initial assessment, we therefore chose the largest and the slowest candidate as the legacy network. The task now is to produce a family of the `backyard dog' networks from this legacy system which fit the imposed hardware constraints and, at the same time, deliver reasonable recognition accuracy.  In the next section we present a technology and an algorithm for creation of  the `backyard dog' networks from a given legacy net.

\section{The `backyard dog' generator}\label{BDresults}

Consider a general legacy network, and suppose that we have access to inputs and outputs for each layer of the network.  Let the input to the first layer be an RGB image. One can now push this input through the first layer and generate this layer's outputs. Output of the first layer become the first-layer features. For a multi-layer network, this process, if repeated throughout the entire network, will define features for each layer. At each layer, these features describe image characteristics that are relevant to the task which the network was trained on.  As a general rule of thumb, as features of the deeper layers show higher degree of robustness. At the same time, these robustness comes at the price of increased memory and computational costs.

\begin{figure}[ht]
\centering\includegraphics[width=0.8\linewidth]{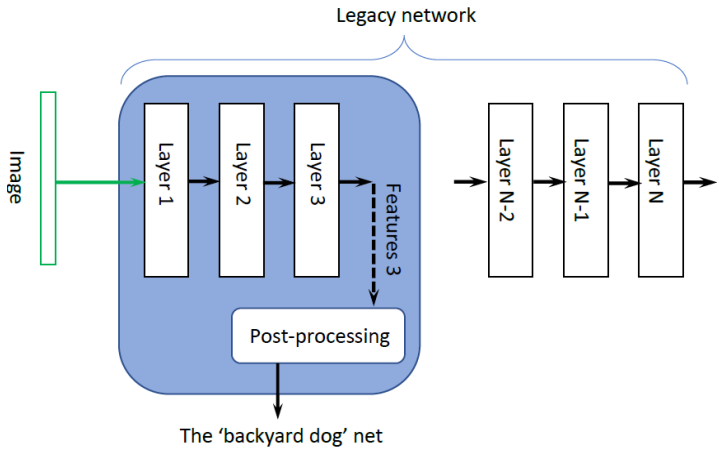}
\caption{Construction of a `backyard dog' network from a given legacy net.} \label{DogNet}
\end{figure}

In our approach, we propose to seek a balance between the requirement of the task at hand, robustness (performance), and computational resources need. To achieve this balance, we suggest to assess suitability of the legacy system's features layer by layer whereby determining the sufficient depth of the network and hence computational resources. The process is illustrated with Fig. \ref{DogNet}. The `backyard dog' net is a truncated legacy system whose outputs are fed into a post-processing routine.

In principle, all layer types could be assessed. In practice, however, it may be beneficial to remove all fully connected layers from the legacy system first. This allows using image scaling as an additional hyper-parameter. This was the approach which we adopted here too.

The post-processing routine itself consisted of several stages:
\begin{itemize}
\item {\it centralization};  subtraction of the mean vector calculated on the training set

\item{\it spherical projection};  projection of the data onto a unit sphere centered at the origin (normalise each data vector to unit length)

\item{ \it construction of new fully connected layer};  the output of this (linear in our case) layer is the output feature vector of the `backyard dog'.
\end{itemize}

\begin{figure}[ht]
\centering\includegraphics[width=0.8\linewidth]{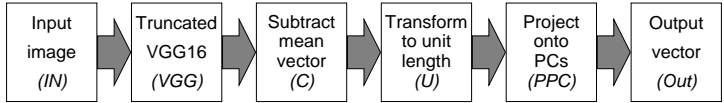}
\caption{Structure of created NN, abbreviations in brackets define notation of corresponding vectors} \label{struct}
\end{figure}

Operational structure of the resulting network is shown in Figure~\ref{struct}. Note that the first processing stage, centralization, can be implemented  as a network layer subtracting a constant vector from the input. The second stage is a well-known $L_2$ normalisation  used, for example, in NN1 and NN2 \cite{Schroff2015}. As for the third stage, several approaches may exist. Here we will use advanced supervised PCA  (cf. \cite{Guoqiang2018}). Details of the calculations used in relevant processing stages as well as interpretation of the `backyard dog' net outputs are is provided in the next section.

\subsection{Interpretation of the `backyard dog' output vector}

Consider a set of identities, $P=\{p_1,\ldots,p_n\}$, where $n$ is the total number of persons in the database.   A set of identities $FM=\{f_1,f_2,\ldots,f_m\}$ forms a family ($m$ is the number of FMs in the family). All identities, which are not elements of FM, are called `other persons' or `strangers'. For each person $f$, $Im(f)$ is the set of images of this person, and $|Im(f)|$ is the number of these images.

For an image $q$, we denote network output as $Out(q)$. Consider:
\begin{equation}
d(q) = \min_{f_i\in FM}\min_{r\in Im(f_i)} \|Out(q)-Out(r)\|.
\end{equation}
Let $t>$ be a decision threshold. If $d(q)>t$ then the image $q$ is interpreted as that of a non-family member (image of a `stranger'). If $d(q)\le t$, then we interpret image $q$ as that of FM $f^*$, where $f^*$
\begin{equation}
f^* = \argmin_{f_i\in FM}\min_{r\in Im(f_i)} \|Out(q)-Out(r)\|
\end{equation}

Three types of errors are considered:
\begin{description}
  \item[MF:] Misclassification of a FM. This error occurs when an image $q$ belongs to a member of the set FM but $Out(q)$ is interpreted as `other person' (a `stranger').
  \item[MO:] Misclassification of a `stranger'. This corresponds to a situation  when an image $q$ does not belong to any of identities from FM but  $Out(q)$  is interpreted as FM.
  \item[MR:] Misrecognition of a FM. This is an error when an  image belongs to a member $f_i$ of the set FM but $Out(q)$ is interpreted as an image of another FM.
\end{description}
Error rates are determined as the fractions of  specific error types during testing (measured in \%). The rate of $\mathbf{ MF+MO}$  is the error rate of the `friend or foe' task.

\subsection{Construction of the `backyard dog' fully connected (linear) layer}

Interpretation rules above  induce requirements for the fully connected linear layer: we need to find an $n$-dimensional subspace $S$ in the space of outputs such that  the distance between projections onto $S$ of the outputs corresponding to images of the same person  is small and the distance between projections onto $S$ of the outputs of images corresponding to different persons is relatively large. This problem has been considered and studied, for example, in  \cite{Zinovyev2000}, \cite{KorenCarmel2004}, \cite{Gorban2009}. Here we follow \cite{Mirkes2016}.
Recall that projection of the vector $x$ onto the subspace defined by orthonormal vectors $\{v^i\}$ is $Vx$, where $V$ is a matrix whose  $i$th rows are $v^i$ ($i=1,\ldots, n$). Select the  target functional in the form:
\begin{equation}\label{probleq}
D_C=D_B-\frac{\alpha}{k}\sum_{i=1}^{k} D_{W_i}\to \max,
\end{equation}
where
\begin{itemize}
\item[]$k$ is  the number of persons in the training set,
\item[]$D_B$ is the mean squared distance between projections of the network output vectors corresponding to different persons:
\begin{equation}
\begin{split}
D_B=&\frac{1}{\sum_{r=1}^{k-1}\sum_{s=r+1}^k |Im(p_r)||Im(p_s)|)}\\
&\times \sum_{r=1}^{k-1}\sum_{s=r+1}^k\sum_{x\in Im(p_r)}\sum_{y\in Im(p_s)} ||Vx-Vy||^2 ,
\end{split}
\end{equation}
\item[]$D_{W_i}$ is the mean squared distance between projections of the network output vectors corresponding to person $p_i$:
\begin{equation}
D_{W_i}=\frac{1}{|Im(p_i)|(|Im(p_i)|-1)} \sum_{x, y \in Im(p_i), \, x\ne y} ||Vx-Vy||^2 ,
\end{equation}
\item[]   parameter $\alpha$ defines the relative cost for the output features corresponding to images of the same person being far apart. In our work, we varied the value of alpha in the interval $[0.1,2]$.
\end{itemize}
The space of the $n$-dimensional linear subspaces of a finite-dimensional space (the Grassmannian manifold) is compact, therefore, the solution of (\ref{probleq}) exists.  The   orthonormal basis of this space (the matrix  $V$) is, by the definition, the set of the first $n$ Advanced Supervised Principal Components (ASPC)  \cite{Mirkes2016}. They are the first $n$ principal axis of the quadratic form defined from  (\ref{probleq}) \cite{KorenCarmel2004}, \cite{Mirkes2016}.

\subsection{Testing protocol}

In our case study we used a database containing 25,402 images of 654 different identities \cite{database2018} (38.84 images per person, on average). First, 327 identities were randomly selected from the database. These identities represented the set $T$ of non-family members. Remaining 327 identities were used to generate sets of family members. We denote these identities as the the set of family members candidates (FMC). Identities from this latter set  with less than 10 images were removed from the set FMC and added to the set $T$ of non-family members. From the set FMC, we  randomly sampled 100 sets of 10 different identities, as examples of FM. We denote these sampled sets of identities as $T_i$, $i=1, \dots,100$. Elements of the set FMC which do not belong to any of the generate sets $T_i$ were removed from the set FMC and added to the set $T$. As a result of this procedure, the set $T$ contained 404 different identities.

For each truncated VGG16 network, and each image $q$ from the training set $T$ we derived output vectors $VGG(q)$ and determined their mean vector $MVGG$
\begin{equation}\label{MVGG}
MVGG=\frac{1}{\sum_{f\in T}|Im(f)|} \sum_{f\in T} \sum_{q\in Im(f)}VGG(q).
\end{equation}
This was used to construct the subtraction layer of which the output was defined as:
\begin{equation}\label{subtrOut}
C(q)=VGG(q)-MVGG.
\end{equation}
Each such vector $C(q)$ was then normalized to unit length.

Next, we determined ASPCs for the set of vectors $C(q)$, $q\in T$  by solving  (\ref{probleq}). The value of $t$ was chosen to minimize the rate of  MF+MO error, for the given test set $T_i$, given value of $\alpha$  and the number of ASPCs. To determine optimal values of $\alpha$ and the number of ASPCs, we derived the mean values of MF, MO,  MR  across all $T_i$:
\begin{equation}\label{averAcc}
\text{MF}=0.01\sum_{i=1}^{100}\text{MF}(T_i), \text{MO}=0.01\sum_{i=1}^{100}\text{MO}(T_i), \text{MR}=0.01\sum_{i=1}^{100}\text{MR}(T_i)
\end{equation}
as well as their maximal values
\begin{equation}\label{maxAcc}
\text{MF}=\max_{i}\text{MF}(T_i), \text{MO}=\max_{i}\text{MO}(T_i), \text{MR}=\max_{i}\text{MR}(T_i).
\end{equation}
For each of these performance metrics, (\ref{averAcc}), (\ref{maxAcc}), we picked the number of ASPCs and the value of $\alpha$ which corresponded to the minimum of the sum MF+MO.

\subsection{Results}

Results of experiments are summarized in Table~\ref{resTabl1} --\ref{resASPCAAver}. Table Table~\ref{resTabl1} shows the amount of time each `backyard dog' network required to process a single image. Tables \ref{resNoPCAMax} --  \ref{resASPCAAver} show performance of  `backyard dog' networks for varying depths (the number of layers).  The best model for networks with 17 layers used 70 ASPCs, and the optimal network with 5 layers used 60 ASPCs. This 5 layer network with 60 ASPCs  processes a single $64\times 64$ image in under 1s on 1 core of Pi. It also demonstrates a reasonable performance, with the MF+MO error rate below 6\%. Note that the maximal value of the MF+MO rate over 100 randomly selected tests $T_i$ is 1.8 times higher than the average for both 17 layer deep and 5 layer deep networks (with optimal number of ASPCs).

\begin{table}
\caption{Time, in seconds, spent on processing of a single image by different `backyard dog' networks, T1 and T2 label two different tests}\label{resTabl1}
{\footnotesize
\begin{tabular}{|c|c|r|r|r|r|r|r|r|r|}
\hline
\multicolumn{1}{|c|}{Image} & \multicolumn{1}{c|}{Layers} & \multicolumn{2}{c|}{ML}  & \multicolumn{2}{c|}{TF Laptop}  & \multicolumn{2}{c|}{TF Pi} & \multicolumn{2}{c|}{C++}\\\cline{3-10}
\multicolumn{1}{|c|}{size} & & \multicolumn{1}{c|}{T1} & \multicolumn{1}{c|}{T2} & \multicolumn{1}{c|}{T1} & \multicolumn{1}{c|}{T2}& \multicolumn{1}{c|}{T1} & \multicolumn{1}{c|}{T2}& \multicolumn{1}{c|}{Laptop} & \multicolumn{1}{c|}{Pi}\\\hline
224&37&0.67&0.72&7.35&7.05& &&&\\\hline
224&35&0.73&0.67& & & &&&\\\hline
224&31&0.62&0.66& & & &&&\\\hline
128&31&0.25&0.24& & & &&&\\\hline
96&31&0.19&0.21&0.96&0.95&17.08&17.31&&\\\hline
64&31&0.07&0.07&0.61&0.64&11.32&11.28&&\\\hline
96&24&0.12&0.13&0.59&0.43&7.44&8.91&&\\\hline
64&24&0.06&0.06&0.35&0.35&7.20&7.27&1.21&5.69\\\hline
64&17&&&&&&&0.81&3.66\\\hline
64&10&&&&&&&0.39&1.61\\\hline
64&05&&&&&&&0.17&0.70\\\hline
\end{tabular}}
\end{table}
\begin{table}
\caption{Error rates for N05, N10, N17, and N24 without PCA improvement. Error rates are evaluated as the maximal numbers of errors for 100 test sets (\ref{maxAcc})}\label{resNoPCAMax}
{\footnotesize
\begin{tabular}{|c|r|r|r|r|}
\hline
Layers &{MR} & {MF} & {MO} & {MF+MO} \\\hline
24&11.00&11.00&0.01&11.01\\\hline
17&14.39&14.39&2.82&17.22\\\hline
10&16.71&16.71&5.86&22.57\\\hline
5&12.58&12.58&2.57&15.14\\\hline
\end{tabular}}
\end{table}
\begin{table}
\caption{Error rates for N05, N10, N17, and N24 without PCA improvement. Error rates are evaluated as the average numbers of errors for 100 randomly selected test sets (\ref{averAcc})}\label{resNoPCAAver}
{\footnotesize
\begin{tabular}{|c|r|r|r|r|}
\hline
Layers & {MR} & {MF} & {MO} & {MF+MO} \\\hline
24&4.16&4.13&1.09&5.22\\\hline
17&7.69&7.65&1.75&9.39\\\hline
10&10.94&10.82&3.64&14.46\\\hline
5&6.58&6.52&2.01&8.53\\\hline
\end{tabular}}
\end{table}
\begin{table}
\caption{Error rates for networks with 5 and 17 layers and optimal number of ASPCs. Error rates are evaluated as the maximal numbers of errors  for 100 randomly selected test sets (\ref{maxAcc})}\label{resASPCAMax}
{\footnotesize
\begin{tabular}{|c|r|r|r|r|}
\hline
Layers & {MR} & {MF} &{MO} & {MF+MO}\\\hline
17&4.80&4.80&1.22&6.02\\\hline
5&9.69&8.16&2.06&10.22\\\hline
\end{tabular}}
\end{table}
\begin{table}
{\footnotesize
\caption{Error rates for networks with 5 and 17 layers and optimal number of ASPCs, errors are evaluated as the average numbers of errors for 100 randomly selected test sets (\ref{averAcc})}\label{resASPCAAver}
\begin{tabular}{|c|r|r|r|r|}
\hline
Layers & {MR} & {MF} &{MO} & {MF+MO}\\\hline
17&2.50&2.46&0.81&3.27\\\hline
5&4.39&4.30&1.48&5.78\\\hline
\end{tabular}}
\end{table}

\section{Conclusion}\label{Conclusion}

In this work we proposed a simple non-iterative method for shallowing down legacy deep convolutional networks. The method is generic in the sense that it applies to a broad class of feed-forward networks, and is based on the Advanced Supervised Principal Component Analysis. We showed that, when applied to the state-of-the-art models developed for face recognition purposes, our approach generates a shallow network with reasonable performance in a specific task. The method enables one to produce of families of smaller-size shallower specialized networks tuned for specific operational conditions and tasks from a single larger and more universal legacy network. The approach and technology were illustrated with a VGG-16 model. They will, however, apply to other models, including the popular MobileNet and SqueezeNet architectures. Testing the approach on these models was beyond the scope and intentions of this paper and will be the subject of our future work.

\section*{Notes}

\subsection*{Acknowledgment}

We are grateful to Prof. Jeremy Levesley for numerous discussions and suggestions in the course of the project.

\subsection*{Funding}

This study was funded by by the {Ministry of Education and Science} of Russia (Project No. 14.Y26.31.0022) and  Innovate UK  Knowledge Transfer Partnership grants KTP009890 and KTP010522.

\subsection*{Compliance with Ethical Standards}

\subsubsection*{Ethical approval}

This article does not contain any studies with human participants performed by any of the authors.

\subsubsection*{Conflict of interests}

The authors declare that they have no conflict of interest.


\begin{thebibliography}{11}



\bibitem{Krizhevsky2012} Krizhevsky, A., Sutskever, I. and Hinton, G.E.. Imagenet classification with deep convolutional neural networks. In Advances in neural information processing systems,  1097--1105, 2012.

\bibitem{Huiying2019} Liang, Huiying, Brian Y. Tsui, Hao Ni, Carolina CS Valentim, Sally L. Baxter, Guangjian Liu, Wenjia Cai et al. Evaluation and accurate diagnoses of pediatric diseases using artificial intelligence. Nature medicine 25 433--438,  2019.

\bibitem{Xiao2019} Sun, Xiao, Lv, Man. Facial Expression Recognition Based on a Hybrid Model Combining Deep and Shallow Features, Cognitive Computation, 2019, https://doi.org/10.1007/s12559-019-09654-y.

\bibitem{Ranjan2018} Ranjan, Rajeev, Swami Sankaranarayanan, Ankan Bansal, Navaneeth Bodla, Jun-Cheng Chen, Vishal M. Patel, Carlos D. Castillo, and Rama Chellappa. Deep learning for understanding faces: Machines may be just as good, or better, than humans. IEEE Signal Processing Magazine 35,  (1):  66--83, 2018.



\bibitem{Zhong-Qiu2019} Zhao, Zhong-Qiu, Peng Zheng, Shou-tao Xu, and Xindong Wu. Object detection with deep learning: A review. IEEE transactions on neural networks and learning systems, 2019.

\bibitem{LeCun2015} LeCun, Yann, Yoshua Bengio, and Geoffrey Hinton. Deep learning. Nature 521(7553) : 436--444, 2015.

\bibitem{Gordienko1993}Gordienko, P. Construction of efficient neural networks: Algorithms and tests. In Neural Networks, 1993. IJCNN'93-Nagoya. Proceedings of 1993 International Joint Conference on 1993 Oct 25 (Vol. 1, pp. 313-316). IEEE.

\bibitem{Gorban1990}Gorban, A.N. Training Neural Networks, USSR-USA JV ``ParaGraph'', 1990.

\bibitem{Hassibi1993}Hassibi, B., Stork, D.G., Wolff, G.J. Optimal brain surgeon and general network pruning. In Neural Networks, 1993., IEEE International Conference on 1993 (pp. 293-299). IEEE.

\bibitem{MobileNet} Howard, Andrew G., Menglong Zhu, Bo Chen, Dmitry Kalenichenko, Weijun Wang, Tobias Weyand, Marco Andreetto, and Hartwig Adam. Mobilenets: Efficient convolutional neural networks for mobile vision applications. arXiv preprint arXiv:1704.04861, 2017.

\bibitem{SqueezeNet} Iandola, Forrest N., Song Han, Matthew W. Moskewicz, Khalid Ashraf, William J. Dally, and Kurt Keutzer. SqueezeNet: AlexNet-level accuracy with 50x fewer parameters and< 0.5 MB model size. arXiv preprint arXiv:1602.07360, 2016.

\bibitem{LiWangKong2017}Li, D., Wang, X., Kong, D. DeepRebirth: Accelerating deep neural network execution on mobile devices. arXiv preprint, 2017 Aug 16 \url{https://arxiv.org/abs/1708.04728}.

\bibitem{EfficientNet} Mingxing, Tan, Quoc V. Le. EfficientNet: Rethinking Model Scaling for Convolutional Neural Networks.
arXiv preprint arXiv:1905.11946, 2019.


\bibitem{Simonyan2015}Simonyan K.,  Zisserman A. Very deep convolutional networks for large-scale image recognition. In International Conference on Learning Representations, 2015.

\bibitem{Parkin2015} Parkhi. O.M., Vedaldi,  A., Zisserman, A. Deep face recognition, Proceedings of the British Machine Vision Conference (BMVC), 2015. \url{http://www.robots.ox.ac.uk/~vgg/publications/2015/Parkhi15/parkhi15.pdf}

\bibitem{Schroff2015} F. Schroff, D. Kalenichenko, and J. Philbin. Facenet: A unified embedding for face recognition and clustering. In Proc. CVPR, 2015.


\bibitem{Taigman2014}Taigman, Y., Yang, M., Ranzato, M.,   Wolf, L. Deep-Face: Closing the gap to human-level performance in face verification. In Proc. CVPR, 2014.

\bibitem{Guoqiang2018} Zhong, Guoqiang, Shoujun Yan, Kaizhu Huang, Yajuan Cai, and Junyu Dong. Reducing and stretching deep convolutional activation features for accurate image classification, Cognitive Computation 10(1): 179--186, 2018.

\bibitem{Mirkes2016}Mirkes, E.M., Gorban A.N., Zinoviev A. Supervised PCA (2016). \url{https://github.com/Mirkes/SupervisedPCA}.

\bibitem{KorenCarmel2004}Koren, Y.,  Carmel, L. Robust linear dimensionality reduction. IEEE Transactions on Visualization and Computer Graphics, 10(4), 459--470, 2004. \url{https://doi.org/10.1109/TVCG.2004.17}


\bibitem{Chen2012}D. Chen, X. Cao, L. Wang, F. Wen, and J. Sun. Bayesian face revisited: A joint formulation. In Proc. ECCV, pages 566–579, 2012.

\bibitem{Sun2014}Sun, Y.,  Wang, X.,  Tang , X.  Deep learning face representation from predicting 10,000 classes. In Proc. CVPR, 2014.


\bibitem{bias2018} Steve Lohr. Face recognition is accurate, if you are a white guy. https://www.nytimes.com/2018/02/09/technology/facial-recognition-race-artificial-intelligence.html, The New York Times, 2018



\bibitem{WhiteErrorRate2015}White, D., Dunn, J.D., Schmid, A.C.,   Kemp, R.I., Error Rates in Users of Automatic Face RecognitiOn Software, PLOS One 10(10): e0139827, 2015 \url{https://doi.org/10.1371/journal.pone.0139827}

\bibitem{Population}Population of the Earth, \url{http://www.worldometers.info/world-population/}

\bibitem{VGG2014}Published VGG CNN \url{http://www.vlfeat.org/matconvnet/models/vgg-face.mat}

\bibitem{MatConvNet}MatConvNet \url{http://www.vlfeat.org/matconvnet}

\bibitem{VGGTF}VGG in TensorFlow \url{https://www.cs.toronto.edu/~frossard/post/vgg16/}



\bibitem{Zinovyev2000}Zinovyev, A.Y. Visualisation of multidimensional data, Krasnoyarsk: Krasnoyarsk State Technocal University Press, 2000 (In Russian).



\bibitem{Gorban2009}Gorban, A.N., Zinovyev, A.Y. Principal Graphs and Manifolds, Chapter 2 in: Handbook of Research on Machine Learning Applications and Trends: Algorithms, Methods, and Techniques, Emilio Soria Olivas et al. (eds), IGI Global, Hershey, PA, USA, 2009, pp. 28-59.


\bibitem{database2018}Gorban A.N., Mirkes E.M., Tyukin I.Y. Preprocessed database LITSO654 for face recognition testing \url{https://drive.google.com/drive/folders/10cu4u-31I24pKTOTIErjmie8gU-Z8biz?usp=sharing}.








\end{thebibliography}
\end{document}